\documentclass{article}

     \usepackage[final, nonatbib]{tackling_climate_workshop_style}

\usepackage[utf8]{inputenc} 
\usepackage[T1]{fontenc}    
\usepackage{hyperref}       
\usepackage{url}            
\usepackage{booktabs}       
\usepackage{amsfonts}       
\usepackage{nicefrac}       
\usepackage{microtype}      
\usepackage{float}          
\usepackage{graphicx}
\usepackage{subcaption}
\usepackage{xcolor}
\usepackage[linesnumbered,ruled,vlined]{algorithm2e}
\usepackage{cite}
\usepackage{tikz}

\hypersetup{colorlinks=true, citecolor=blue}

\SetCommentSty{mycommfont}

\title{Adaptive-Labeling for Enhancing Remote Sensing Cloud Understanding}

\usepackage{authblk}
\makeatletter
\renewcommand\AB@affilsepx{, \protect\Affilfont}
\makeatother
\begin{document}
\author[1]{\textbf{Jay Gala}}
\author[2]{\textbf{Sauradip Nag}}
\author[3]{\textbf{Huichou Huang}}
\author[4]{\textbf{Ruirui Liu}}
\author[2]{\textbf{Xiatian Zhu}}
\affil[1]{NMIMS University}
\affil[2]{University of Surrey}
\affil[3]{City University of Hong Kong}
\affil[4]{Brunel University London}
\maketitle

\begin{abstract}
Cloud analysis is a critical component of weather and climate science, impacting various sectors like disaster management. However, achieving fine-grained cloud analysis, such as cloud segmentation, in remote sensing remains challenging due to the inherent difficulties in obtaining accurate labels, leading to significant labeling errors in training data.
Existing methods often assume the availability of reliable segmentation annotations, limiting their overall performance. To address this inherent limitation, we introduce an innovative model-agnostic {\em Cloud Adaptive-Labeling} ({\bf CAL}) approach, which operates iteratively to enhance the quality of training data annotations and consequently improve the performance of the learned model.
Our methodology commences by training a cloud segmentation model using the original annotations. 
Subsequently, it introduces a {\em trainable pixel intensity threshold} for adaptively labeling the cloud training images on the fly. The newly generated labels are then employed to fine-tune the model.
Extensive experiments conducted on multiple standard cloud segmentation benchmarks demonstrate the effectiveness of our approach in significantly boosting the performance of existing segmentation models. Our CAL method establishes new state-of-the-art results when compared to a wide array of existing alternatives.


\end{abstract}

\section{Introduction}

Cloud understanding is indicative and critical in climate science, with significant application in various sectors like renewable energy, disaster management, and environmental monitoring \cite{li-2016}.
A viable and scalable approach to this problem is leveraging remote sensing imagery data
to examine the patterns and characteristics of clouds over space and time.
To that end, cloud segmentation is one of the most fine-grained methods.

Traditionally, cloud segmentation primarily employed threshold-based techniques, hinging on pre-existing knowledge and the distinction between Earth's surface and clouds \cite{rossow-1993}. Nevertheless, these methods exhibited vulnerability to intricate backgrounds, thereby constraining their efficacy. Additionally, the selection of the threshold in such methods typically entailed subjective human judgment and expertise, resulting in performance disparities \cite{li-2021}.




Recently, deep learning approaches have emerged as promising alternatives to threshold-based methods. They excel in capturing intricate patterns and leveraging extensive datasets, resulting in significant enhancements in cloud segmentation accuracy \cite{luo-2022}. Deep learning mitigates the reliance on subjective thresholding while adapting effectively to diverse atmospheric conditions and complex cloud formations \cite{ge-2023}.

\begin{figure}
  \begin{subfigure}{0.22\textwidth}
    \centering
    \begin{tikzpicture}[baseline=(current bounding box.center)]
      \node[draw=black, line width=0.2pt, inner sep=1pt] {\includegraphics[width=\linewidth]{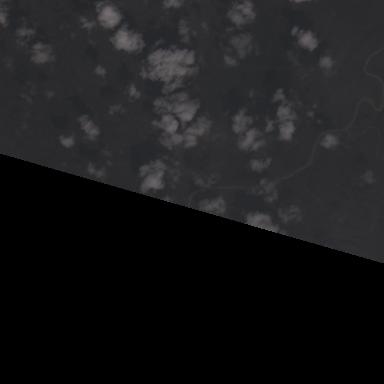}};
    \end{tikzpicture}
    \caption{}
  \end{subfigure}
  \hfill
  \begin{subfigure}{0.22\textwidth}
    \centering
    \begin{tikzpicture}[baseline=(current bounding box.center)]
      \node[draw=black, line width=0.4pt, inner sep=1pt] {\includegraphics[width=\linewidth]{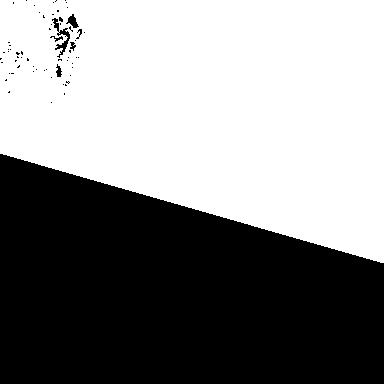}};
    \end{tikzpicture}
    \caption{}
  \end{subfigure}
  \hfill
  \begin{subfigure}{0.22\textwidth}
    \centering
    \begin{tikzpicture}[baseline=(current bounding box.center)]
      \node[draw=black, line width=0.2pt, inner sep=1pt] {\includegraphics[width=\linewidth]{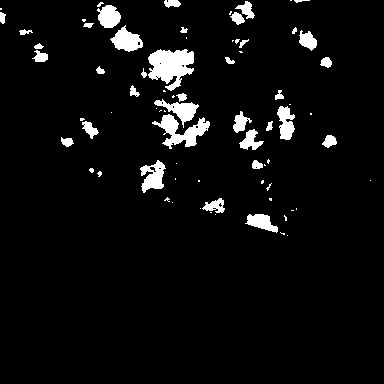}};
    \end{tikzpicture}
    \caption{}
  \end{subfigure}
  \caption{Illustration of noisy cloud mask labeling. (a) A random training image; (b) The original cloud mask label; (c) The mask label obtained using our proposed method.}
  \label{fig:three_images}
\end{figure}

Nonetheless, all the previous methods typically assume the availability of accurate mask annotations of cloud, which is often largely invalid (see Figure \ref{fig:three_images}(a)). Making this assumption could generally lead to degraded model performance, but it is still understudied in the literature.


In this paper, we propose an innovative model-agnostic {\em Cloud Adaptive-Labeling} ({\bf CAL}) method to address the mentioned fundamental challenge. Our approach starts by training a cloud segmentation model using the original noisy training data. Subsequently, it introduces a {\em trainable pixel intensity threshold} for adaptively labeling the training images on the fly. The newly generated labels are then employed to fine-tune the model.
Importantly, our pixel intensity threshold is dynamically adjusted through an intuitive self-regulating mechanism in response to the evolving training loss: it increases when the loss decreases and decreases when it increases. 
This objective-oriented thresholding strategy challenges the conventional notion of a single optimal threshold and obviates the necessity for prior threshold estimation.
Extensive experiments demonstrate that our method iteratively enhances the quality of training data annotations, leading to improvements in model performance and achieving the new state of the art on a standard cloud segmentation benchmark.




\section{Related work}


Existing cloud segmentation techniques are categorized into three groups: Threshold-based, Machine learning-based, and Deep learning-based. Threshold-based methods rely on differences in scattering intensity between clouds and atmospheric particles in specific spectral bands \cite{heinle-2010}. 
The spectral threshold approach is effective when combined with multi-band features like radiance, reflectance, and normalized vegetation index \cite{frantz-2018,huang-2010,zhu-2012}. These methods work well only when clouds have high visible reflection and cooler infrared temperatures compared to the Earth's surface. However, their performance depends on prior knowledge and the contrast between clouds and the Earth's surface. Despite their effectiveness, threshold-based methods struggle to distinguish clouds in the presence of atmospheric aerosols, dust, smoke, and hazy pixels caused by small-scale cumulus clouds or cloud edges \cite{li-2021}.


To address the limitations of threshold-based approaches, following-up works employed machine learning algorithms, such as Support Vector Machines (SVMs) and random forests \cite{zhu2013video,zhu2014constructing}, leveraging texture features extracted from remote sensing images as input \cite{sui-2019, wei-2020}. However, these methods heavily depended on manually crafted features. 
Recent advancements have witnessed the adoption of deep learning techniques in cloud segmentation, with subsequent efforts focused on enhancing network architecture and deep feature extraction \cite{chai-2019, xie-2017, yang-2019}. For instance, Yao et al. \cite{yao-2021} introduced attention modules and boundary refinement blocks to comprehensively capture multi-scale context.

While deep learning-based cloud detection methods exhibit significant advantages over threshold-based and traditional machine learning approaches, their outstanding performance hinges on the availability of extensive, high-quality pixel-level cloud masks \cite{francis-2019}. Nevertheless, the creation of such datasets is a challenging and time-consuming endeavor, often resulting in the presence of substantial noisy labels that can introduce uncertainties and negatives into research outcomes (see Figure \ref{fig:three_images}). 
Interestingly, this problem has never been systematically considered and investigated in the literature. 
This study marks the inaugural endeavor to address the challenge of noisy cloud mask labeling by introducing a model-agnostic CAL method within the deep learning domain.

\section{Cloud Adaptive Labeling}

  
  

\begin{figure}
\begin{center}
  \begin{subfigure}{0.99\textwidth}
    \centering
    \includegraphics[width=\linewidth]{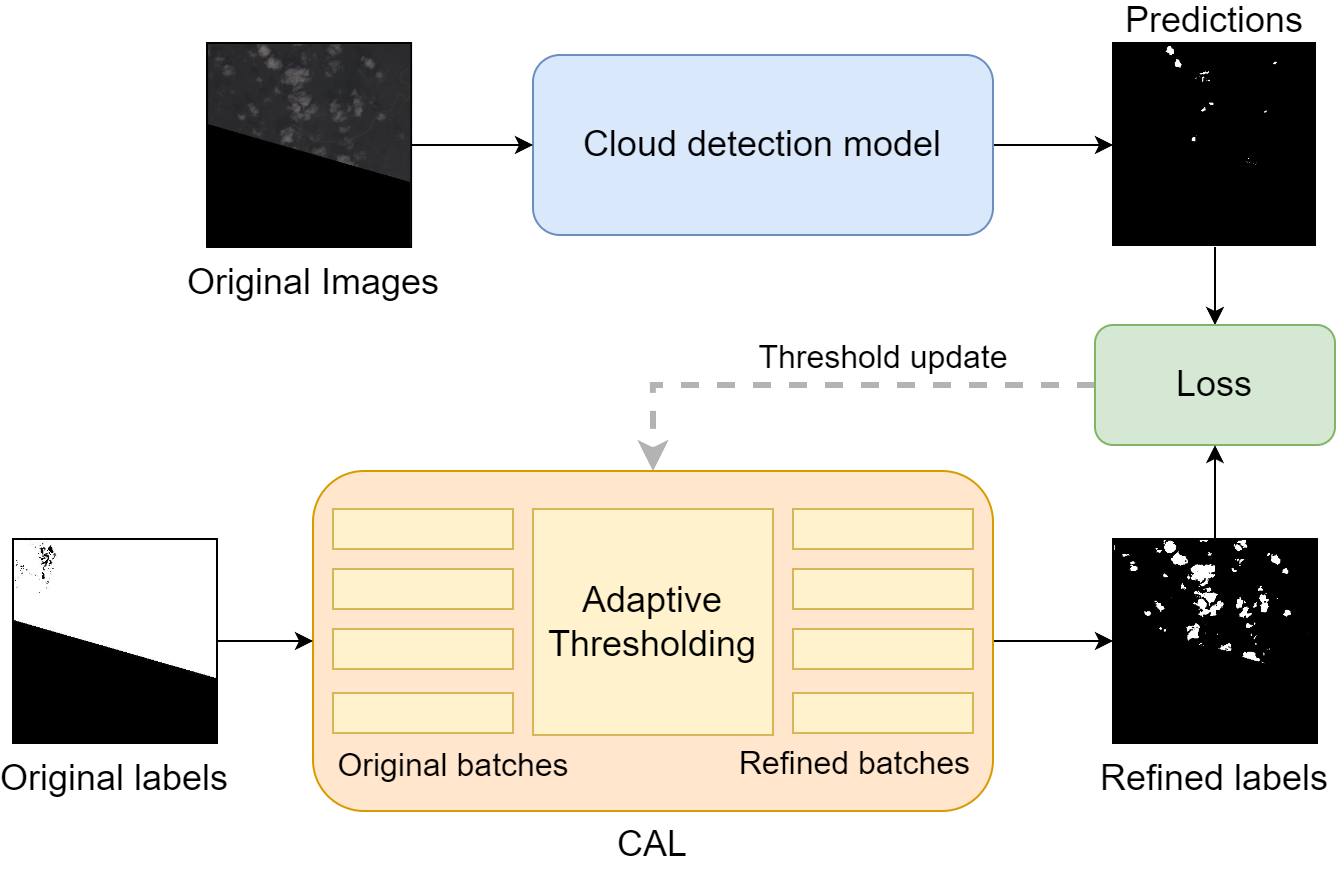}
    \caption{}
  \end{subfigure}

  \begin{subfigure}{0.99\textwidth}
    \centering
    \includegraphics[width=\linewidth]{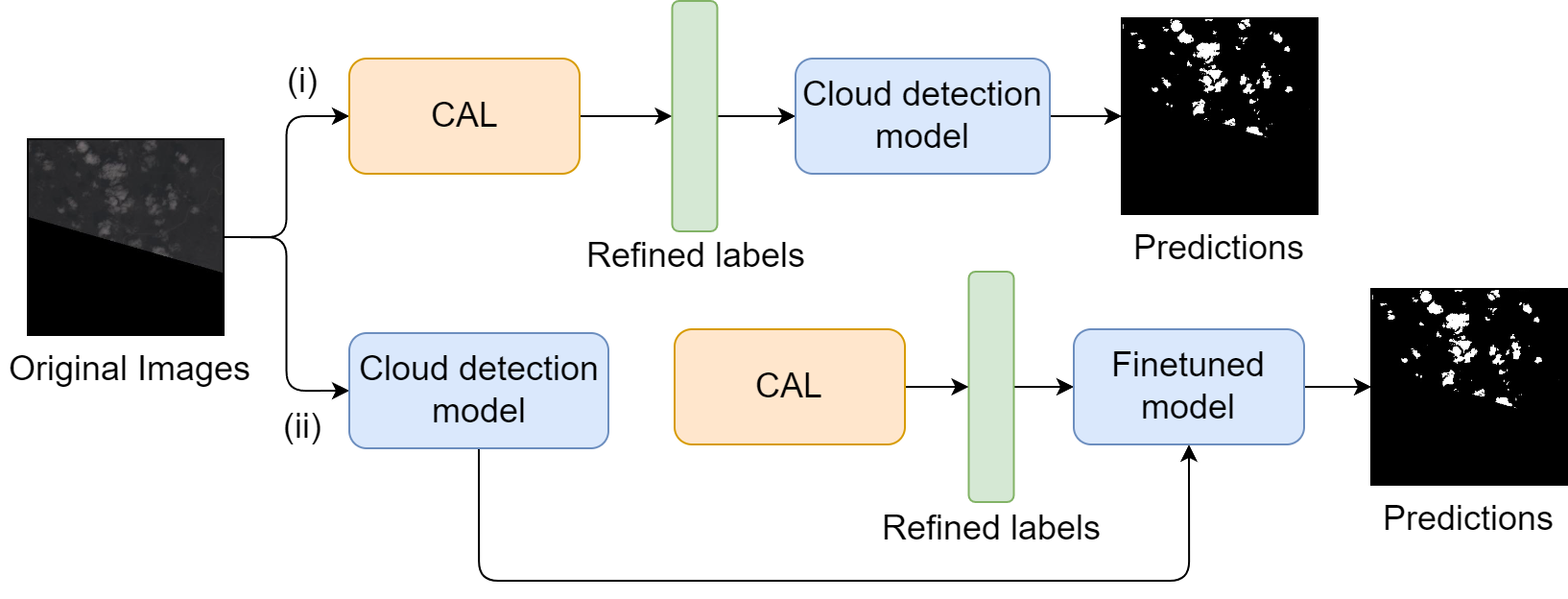}
    \caption{}
  \end{subfigure}
  \caption{(a) \textbf{Overview for our proposed {\em Cloud Adaptive-Labeling} (CAL) method.}
  Our approach begins with training a cloud segmentation model using the original training data, which inherently contains noise. We then incorporate a trainable pixel intensity threshold to adaptively label the cloud mask by applying this threshold to the input training images. These newly generated cloud mask labels are subsequently used for fine-tuning the model. Importantly, our threshold is dynamically adjusted based on the training loss: it increases when the loss decreases and decreases when the loss increases. This innovative training objective-driven thresholding strategy not only challenges the traditional assumption of a single optimal threshold but also eliminates the need for prior estimation of an appropriate threshold. (b) \textbf{Integrating CAL} (i) during finetuning or (ii) after finetuning}
  \label{fig:model overview and integration}
  \end{center}
\end{figure}

{\bf Overview.} In contrast to the previous methods, which prioritized improving the model's efficiency and its ability to capture the relationships between the images and their labels, we propose a method to reduce the underlying noise in the cloud segmentation datasets. In this section, we describe CAL, which comprises an efficient 
module that enhances the detection capability of existing cloud detection approaches. It is normally plugged in as a post-processing block which relabels the existing noisy cloud labels into a refined version and when integrated into the training scheme boosts the performance of the existing pretrained model. As depicted in Fig. \ref{fig:model overview and integration}(a), it is mainly composed of the novel dynamic thresholding module that is adaptive with respect to the noise in the labels.

{\bf Problem statement.} Given a satellite image $I$, the task of cloud detection is to predict a pixel-level binary classification of whether it is a cloud or not and finally outputs a binary mask $M$. Formally, given a training dataset $D_{train}$, it is composed of an image $I$ and masked label $M$ pair which is used for training the cloud detection model and then it is inferred using a testing split $D_{test}$.

{\bf CAL algorithm.}
Serving as a model-agnostic plug-in,
our CAL is designed to gradually improve mask labels during training, boosting model performance.
We initialize a per-mini-batch threshold parameter $p_{th}$, which learns based on changes in model predictions.
We use $p_{th}$	to binarize the input images as follows: 
\begin{equation}
    M_{cal} = \phi(I, p_{th})
\end{equation}
where $\phi(\cdot)$ is the morphological image binarization operation.
We replace the original/previous mask labels with $M_{cal}$, which has better quality.
We then further fine-tune the model with the new labels using the same objective function as the baseline detection model, e.g., binary cross-entropy loss \cite{zhang-2018}.


{\bf\em Threshold adaptation}: 
We dynamically adjust the threshold $p_{th}$ according to the objective loss value of the new binary mask $M_{cal}$. If the loss decreases, we decrease $p_{th}$ by a small preset amount and vice versa. See Algorithm \ref{algo} for details.

\begin{algorithm}[!ht]
\DontPrintSemicolon
  
  
  \BlankLine
  \texttt{learnable\_threshold = 60.0}\ \tcp*{initial threshold}
  \texttt{delta\_x = 2.0}\ \tcp*{step size}
  \texttt{best\_loss = float('inf')}\;
  \texttt{lower\_bound = 45.0}\;
  \texttt{update\_frequency = 150} 
  
  \BlankLine
    \While{\texttt{not done}}{
      \texttt{new\_labels = binarize\_image(images, threshold=learnable\_threshold)}\;
      \texttt{outputs = model(images)}\;
      \texttt{loss = criterion(outputs, new\_labels)}\;
  
        \texttt{best\_loss = min(best\_loss, loss)}\;
  
      \If{\texttt{(idx + 1) \% update\_frequency == 0}}{
        \If{\texttt{loss > best\_loss}}{
          \texttt{learnable\_threshold -= delta\_x} \tcp*{Decrease the threshold}
          \texttt{learnable\_threshold = max(lower\_bound, learnable\_threshold)}\;
        }
        \Else{
          \texttt{learnable\_threshold += delta\_x} \tcp*{Increase the threshold}
        }
      }
    }
\caption{Cloud Adaptive Labeling}
\label{algo}
\end{algorithm}

{\bf Integrating CAL with existing methods.} CAL can be integrated with existing cloud detection models without altering the design or adding any learnable parameters (see Figure \ref{fig:model overview and integration}(b)). CAL can be applied directly to any existing model without retraining. However, retraining with CAL has demonstrated enhanced results.


\section{Experiments}

{\bf Dataset.}
In our experiments, we utilize the publicly accessible 38-Cloud dataset
\cite{mohajerani-2022}. 
This dataset comprises 38 Landsat-8 scenes, each with dimensions of 1000$\times$1000 pixels, and includes manually generated pixel-level cloud mask ground truths, albeit with some noise. The dataset is structured into numerous patches, each measuring 384$\times$384 pixels, and is composed of four spectral channels: Red, Green, Blue, and Near Infrared (NIR). Specifically, there are 8,400 patches designated for training purposes and an additional 9,201 patches allocated for testing.




{\bf Implementation details.}
We trained our CAL model using the Adam optimizer \cite{kingma-2014} with a learning rate of 0.001 for 3 epochs. 
The initial threshold value was set to 60.0, with a step size of 2.0. We defined the best loss as infinity and the lower bound as 45.0. Additionally, we updated the threshold every 150 steps. As an illustration, we integrated CAL with the widely used U-Net \cite{ronneberger-2015}, DeeplabV3 \cite{chen-2017}, and FCN \cite{shelhamer-2017} segmentation models.

{\bf Results.}
Table \ref{38-cloud} reveals the following insights:
(1) Our baseline models, U-Net, DeeplabV3, and FCN, perform notably worse than specialized methods like LWCDnet. This discrepancy suggests the presence of unique challenges in cloud segmentation.
(2) Encouragingly, our CAL approach significantly enhances the performance of U-Net, DeeplabV3, and FCN, demonstrating an improvement of 20\% or more in mIoU. This achievement positions our method as the top-performing solution among all competitors, underscoring its effectiveness in mitigating the challenges posed by noisy cloud labeling.
In particular, U-Net+CAL achieves the best performance among all the competitors. (3) In addition to this, our proposed CAL consistently improves the performance of FCN, DeeplabV3, and U-Net-based cloud detection models by simply plugging in without any design changes, thus making our algorithm model agnostic in nature.

\begin{table}[H]

  \caption{Performance comparison on the 38-Cloud dataset.}
  \label{38-cloud}
  \centering
  \footnotesize
  \begin{tabular}{lccccc}
    \toprule
    Method     &   mIoU     &  Precision & Recall & F1-Score & OA \\
    \midrule
    CD-CTFM \cite{ge-2023} & 84.13  & 91.09 & 89.22     & 90.15 & 95.45 \\ 
    CD-AttDLV3+ \cite{yao-2021}& 81.24  & 88.85 & 87.58     & 88.21 & 94.49 \\
    CloudAttU \cite{guo-2020} & 84.73 & 90.62 & 89.95     & 90.28 & 95.92 \\
    CloudFCN \cite{francis-2019} & 83.31 & 88.81 & 89.61     & 89.21 & 95.66 \\
    DeeplabV3+ \cite{chen-2017} & 81.64 & 87.72 & 88.36 & 88.04 & 95.03\\
    CD-Net \cite{yang-2019} & 89.70 & 94.30 & 94.70 & 94.50 & 95.40\\
    LWCDnet \cite{luo-2022} & 89.90 & 95.10 & 94.30 & 94.70 & 95.30\\
    FCN \cite{shelhamer-2017} & 60.40  & 80.69 & 63.02 & 69.07 & 92.38 \\  
    DeeplabV3 \cite{chen-2017} & 44.02 & 86.25 & 89.53 & 86.73 & 96.07 \\
    U-Net \cite{ronneberger-2015} & 73.69  & 74.31 & 97.17     & 82.10 & 91.20 \\ 
    \midrule
    FCN + \bf CAL & 83.94 & 91.52 & 87.73 & 89.26 & 96.74\\
    DeeplabV3 + \bf CAL & 83.95 & 85.95 & 94.17 & 89.38 & 96.77\\
    \textbf{U-Net + CAL} & \textbf{93.15} & \textbf{96.37} & \textbf{96.67} & \textbf{96.44} & \textbf{98.61}\\
    
    \bottomrule
  \end{tabular}
\end{table}

{\bf Model agnosticity of CAL.} Our proposed CAL works as a plug-and-play module on existing cloud detection models like FCN, Deeplabv3, and U-Net respectively. From the results in Table \ref{ablation-all}, it is seen that our CAL when plugged in improves consistently over the baseline model without CAL. This is observed for all the models thus proving the model-agnosticity of our algorithm design. 

\begin{table}[H]
  \caption{Performance comparisons of models with and without CAL.}
  \label{ablation-all}
  \centering
  \begin{tabular}{lccccc}
    \toprule
    Methods     &   mIoU     &  Precision & Recall & F1-Score & OA \\
    \midrule
    FCN & 60.40  & 80.69 & 63.02 & 69.07 & 92.38 \\ 
    FCN + CAL (ours)& 83.94 & 91.52 & 87.73 & 89.26 & 96.74\\
    \midrule
    U-Net & 73.69  & 74.31 & 97.17     & 82.10 & 91.20 \\
    U-Net + CAL (ours)& 93.15 & 96.37 & 96.67 & 96.44 & 98.61\\
    \midrule
    DeeplabV3 & 44.02  & 86.25 & 89.53 & 86.72 & 96.07 \\     
    DeeplabV3 + CAL (ours) & 83.95 & 85.95 & 94.18 & 89.39 & 96.77\\
    \bottomrule
  \end{tabular}
\end{table}

\begin{figure}
    \centering
    \begin{subfigure}{0.45\textwidth}
        \includegraphics[width=\linewidth]{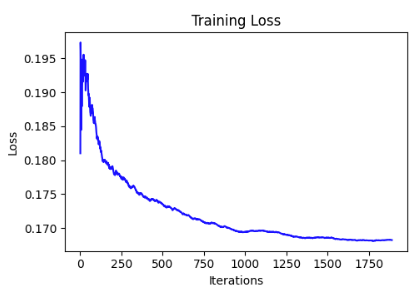}
        \caption{}
    \end{subfigure}
    \begin{subfigure}{0.45\textwidth}
        \includegraphics[width=\linewidth]{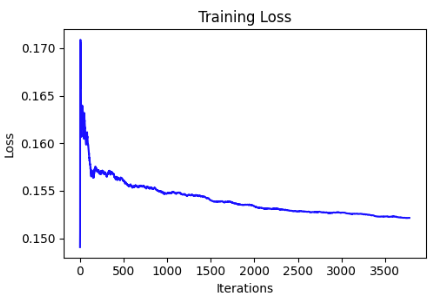}
        \caption{}
    \end{subfigure}

    \begin{subfigure}{0.45\textwidth}
        \includegraphics[width=\linewidth]{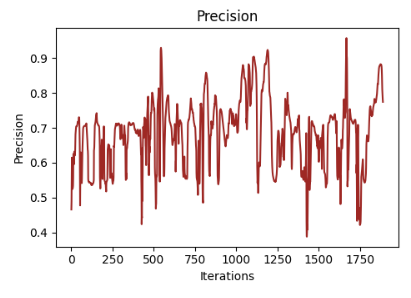}
        \caption{}
    \end{subfigure}
    \begin{subfigure}{0.48\textwidth}
        \includegraphics[width=\linewidth]{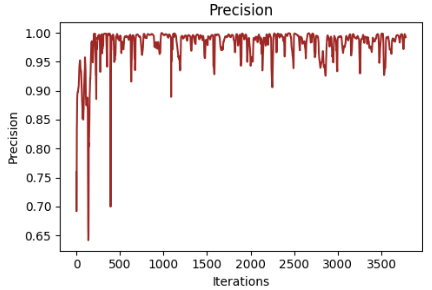}
        \caption{}
    \end{subfigure}

    \caption{(a) Training loss curve for finetuning U-Net (b) Training loss curve for fine-tuning U-Net with CAL. Finetuning with CAL results in faster convergence than fine-tuning without CAL (c) Precision curve for finetuning U-Net (d) Precision curve for fine-tuning U-Net with CAL. Finetuning with CAL quickly stabilizes the precision as opposed to fine-tuning without CAL.}
    \label{convergence}
\end{figure}

{\bf Training convergence due to CAL.} Existing labels provided by the model often have incorrect cloud segmentation annotation as shown in Fig \ref{fig:three_images}. Such faulty annotations make the model slower to converge. As seen from Fig \ref{convergence}, the variant with CAL (Fig \ref{convergence} (b)) converges faster than the variant without CAL (Fig \ref{convergence} (a)). This shows the efficacy of our adaptive thresholding improving the labels. Additionally, our CAL improves the stability and accuracy of the model precision than the baseline variant as shown from Fig \ref{convergence} (c,d) respectively.

\section{Conclusion}
In this study, we tackle a key challenge associated with noisy training data in cloud analysis, specifically focusing on cloud segmentation in remote sensing. We propose a novel model-agnostic self-labeling approach that iteratively enhances the quality of training data annotations, resulting in improved cloud segmentation model performance. Our method is designed to seamlessly integrate with various existing segmentation approaches. Extensive experiments conducted on a widely recognized cloud segmentation benchmark validate the effectiveness of our approach. Our method outperforms existing methods, establishing a new state-of-the-art performance level.

\bibliographystyle{unsrt} 
\bibliography{ref}

\end{document}